\documentclass{article} 
\usepackage{iclr2026_conference,times}

\usepackage{hyperref}
\usepackage{url}

\usepackage{booktabs} 

\usepackage{amsmath,amsfonts,bm}

\def\eqref#1{equation~\ref{#1}}

\def\1{\bm{1}}

\DeclareMathAlphabet{\mathsfit}{\encodingdefault}{\sfdefault}{m}{sl}
\SetMathAlphabet{\mathsfit}{bold}{\encodingdefault}{\sfdefault}{bx}{n}

\usepackage{multicol}
\usepackage{adjustbox}

\usepackage{graphicx}
\usepackage{float}
\usepackage[table]{xcolor}
\usepackage{array}
\usepackage{multirow}
\usepackage{makecell}
\usepackage[margin=1in]{geometry}
\fancyheadoffset[LE,RO]{0pt}
\fancyheadoffset[LO,RE]{0pt}
\usepackage{enumitem}
\usepackage[most]{tcolorbox}
\usepackage{lmodern} 
\usepackage[T1]{fontenc}

\usepackage{booktabs}
\usepackage{tabularx}
\usepackage{siunitx}
\usepackage{caption}
\usepackage{ragged2e}
\usepackage[table]{xcolor}
\usepackage{threeparttable}
\usepackage{amsmath}
\usepackage{amssymb}

\usepackage{algpseudocode}
\usepackage{amssymb}
\usepackage{enumitem}
\usepackage[most]{tcolorbox} 

\usepackage{makecell,tikz}
\usepackage{mdframed}

\usepackage{xcolor}
\definecolor{lstframe}{gray}{0.70}   
\definecolor{lstbg}{gray}{0.97}      

\usepackage{makecell,threeparttable}

\usepackage{fancyvrb}

\usepackage{listings}

\colorlet{TitleBG}{black!0}   
\colorlet{MetaHdr}{black!10}  
\colorlet{StatHdr}{black!6}   

\usepackage{algorithm}
\usepackage{algorithmicx}
\usepackage{algpseudocode}

\newcolumntype{L}[1]{>{\raggedright\arraybackslash}p{#1}}
\newcolumntype{C}[1]{>{\centering\arraybackslash}p{#1}}

\newcolumntype{L}[1]{>{\raggedright\arraybackslash}p{#1}}
\newcolumntype{C}[1]{>{\centering\arraybackslash}p{#1}}
\newcolumntype{R}[1]{>{\raggedleft\arraybackslash}p{#1}}

\newcommand{\model}[1]{\texttt{#1}}

\title{Individualized Cognitive Simulation in Large Language Models: Evaluating Different Cognitive Representation Methods}

\author{\noindent
  \textbf{Tianyi Zhang\textsuperscript{1}, Xiaolin Zhou\textsuperscript{1}, Yunzhe Wang\textsuperscript{1}, Erik Cambria\textsuperscript{2}, David Traum\textsuperscript{3}, Rui Mao\textsuperscript{2}} \\
  \textsuperscript{1}Department of Computer Science, University of Southern California, USA\\
  \textsuperscript{2}College of Computing and Data Science, Nanyang Technological University, Singapore \\
  \textsuperscript{3}USC Institute for Creative Technologies \\
  \texttt{\{tzhang62,xzhou733,yunzhewa\}@usc.edu}; 
  \texttt{\{cambria,rui.mao\}@ntu.edu.sg}; \texttt{traum@ict.usc.edu}
}

\iclrfinalcopy 

\begin{document}

\maketitle

\begin{abstract}


Individualized cognitive simulation (ICS) aims to build computational models that approximate the thought processes of specific individuals. While large language models (LLMs) convincingly mimic surface-level human behavior such as role-play, their ability to simulate deeper individualized cognitive processes remains poorly understood. To address this gap, we introduce a novel task that evaluates different cognitive representation methods in ICS. We construct a dataset from recently published novels (later than the release date of the tested LLMs) and propose an 11-condition cognitive evaluation framework to benchmark seven off-the-shelf LLMs in the context of authorial style emulation. We hypothesize that effective cognitive representations can help LLMs generate storytelling that better mirrors the original author. Thus, we test different cognitive representations, e.g., linguistic features, concept mappings, and profile-based information. Results show that combining conceptual and linguistic features is particularly effective in ICS, outperforming static profile-based cues in overall evaluation. Importantly, LLMs are more effective at mimicking linguistic style than narrative structure, underscoring their limits in deeper cognitive simulation. These findings provide a foundation for developing AI systems that adapt to individual ways of thinking and expression, advancing more personalized and human-aligned creative technologies. 
\end{abstract}

\section{Introduction}


Artificial intelligence (AI) was initially conceived as an attempt to replicate human intelligence, drawing inspiration from the mechanisms of human thought and reasoning~\citep{cambria2023seven}. Recently, the rapid development of large language models (LLMs) has marked a qualitative leap in the processing capabilities of AI systems, enabling machines to perform complex linguistic, analytical, and generative tasks with unprecedented fluency and scale. Yet, the broader aspiration of modeling human cognition remains only partially realized. 

Conventional role-play techniques within LLMs have shown promise in imitating professional roles, social identities, and behavioral archetypes~\citep{shanahan2023role}, whereas these simulations are inherently constrained: they reproduce externalized patterns of behavior rather than capturing the subtle, dynamic, and individualized nature of cognition. Extending such methods to represent individual cognitive profiles is difficult because our theoretical and computational understanding of how to represent individual cognition, as distinct from generalized occupational categories such as lawyers, doctors, or researchers, is still limited. The significance of cognitive representations for individuals lies in their potential to uncover the unique ways in which individuals perceive, reason, and act across diverse contexts. It is critical in designing systems that move beyond generic behavioral templates to support adaptive, personalized interactions, whether in education, healthcare, decision-making support, or social simulation~\citep{maocom}.

In this work, we aim to test the utility of different cognitive representation methods in the context of individualized cognitive simulation (ICS). Objectively evaluating cognitive patterns is difficult because cognition itself is multifaceted, context-dependent, and often expressed implicitly through behavior, language, and decision-making rather than in directly observable forms. In light of this, we propose a novel evaluation task that measures how well LLM continuations match the authorial style of original texts after injecting different cognitive representations. Our hypothesis is that prompting LLMs with effective cognitive representations can lead to greater stylistic consistency in storytelling, making their outputs more closely resemble the original authors. This hypothesis is in line with ``\textit{Bayesian Models of Cognition}'' from~\citet{griffiths2001bayesian}, where human behavior is characterized as a conditional distribution: 
\[
P(\text{response} \mid \text{stimulus}, \text{cognitive model})
\]
This formulation links observable responses to external stimuli through the mediating role of internal cognitive structures. By analogy, LLM outputs can be viewed as responses conditioned not only on direct context stimuli but also on injected cognitive representations. Thus, we construct a dataset from recently published novels and propose an 11-condition cognitive evaluation framework to benchmark seven off-the-shelf LLMs in the context of authorial style emulation. To prevent the risk of data leakage~\citep{wu2025antileak}, we source novels that were published later than the release date of the tested LLMs.

The evaluated cognitive representation methods include linguistic features, concept mappings, and author profile information (persona, background, Big Five/OCEAN personality). Linguistic features reflect the direct writing habits of an author; concept mappings capture personalized understandings of abstract concepts and thinking frameworks; and profile information represents prior experiences that may shape unique cognitive, emotional, and stylistic patterns. Together, these dimensions provide complementary perspectives on how cognition manifests in text. We prioritize these representations because of their theoretical grounding and non-invasive accessibility. They are readily derived from text or public reports, making them practical to implement for uncovering the cognitive signatures of individual authors in real-world applications.

By testing on the continuous writing tasks from five authors, we find that the combination of concept mappings and linguistic features achieves the highest overall performance, significantly outperforming both single-feature and other multi-feature settings. This synergy suggests that while linguistic cues capture surface-level stylistic habits, concept mappings provide deeper insight into how authors conceptualize and structure meaning across narrative contexts, and together they offer a complementary pathway to more faithful style emulation. In contrast, profile-based features such as persona, background, or personality traits yield limited improvements and sometimes even degrade performance when combined, highlighting the challenges of translating high-level biographical information into effective narrative generation signals.

\noindent The contribution of this work can be summarized as follows:  1) \textbf{Individualized Cognitive Simulation:} A new task using computational models to approximate individual cognitive and stylistic processes. 2) \textbf{Cognitive Evaluation Framework:} An 11-condition framework that integrates linguistic, conceptual, and profile-based representations into narrative continuation, combining LLM-based automatic metrics and human evaluation to assess the utilities of different cognitive representations. 3) \textbf{Empirical Research Findings:} The combination of concept mappings and linguistic features is the most effective signal for ICS, achieving the strongest overall performance; LLMs mimic linguistic style more effectively than narrative structure, revealing the limits of their deeper cognitive simulation.

\section{Related Work}

Research on ICS draws from computational linguistics, cognitive modeling, and narrative generation. Stylometry and authorship attribution showed how linguistic patterns reveal individual traits~\citep{stamatatos2009survey}, later extended to style transfer and personalized generation with neural models~\citep{shen2017style, fu2018style}. Recent LLM studies explored author imitation and persona-driven writing~\citep{zhang2018personalizing, huang2023lapdog, bhandarkar2024emulating}, which we extend by integrating stylistic, conceptual, and profile signals. Role-play has emerged as another common approach for guiding LLMs to simulate personas or identities. This is typically realized through a short role preamble that assigns the model an identity (e.g., “You are a doctor”)~\citep{dearaujo2024helpful}, in-context exemplars that script how the role should sound or behave~\citep{rupprecht2025rags,zhang2025maps}, and conversational updating that incrementally refines the role as interaction unfolds across turns~\citep{xu2022cosplay,wang2025coser}. However, role-play remains limited to surface behaviors: it can mimic style but struggles to represent the dynamic and individualized cognitive traits of real authors. Recent studies confirm this limitation, with \citet{hu2024quantifying} showing that current persona effects in LLMs are shallow and \citet{pal2025beyond} finding persistent inconsistencies in character and profile fidelity. On the other hand, evaluating generated stories also remains difficult. Prior work has compared human and automatic metrics~\citep{clark2018creative, ippolito2020automatic} and, more recently, tested LLMs as judges~\citep{chiang-lee-2023-large, chang2023survey}. While scaling improves benchmarks~\citep{kaplan2020scaling, wei2022emergent} and decoding affects fluency~\citep{holtzman2020degeneration}, our results suggest that scale alone is insufficient for deeper cognitive simulation.


\section{Preliminary}
\label{sec:preliminaries}
As~\citet{wittgenstein1922} argued, \textit{``the limits of my language mean the limits of my world''}, highlighting the significance of language in reflecting cognition. ICS builds on this idea by treating an author’s narrative patterns as the product of multiple layers of cognitive processes. From a cognitive science perspective, language is not just a sequence of words but a reflection of deeper mental structures, including habitual linguistic choices, conceptual systems, and biographical experiences~\citep{boroditsky2001language}. To approximate these processes computationally, we evaluate three families of cognitive representations in this work, namely linguistic features, concept mappings, and author profiles.

\noindent \textbf{Linguistic features} capture surface-level stylistic patterns that characterize an author’s habitual use of language\citep{crystal1969investigating}. In our framework, these include lexical style (frequent vocabulary and recurring word combinations), syntactic style (part-of-speech distributions, sentence length, and grammatical preferences), semantic themes (recurring topical words that reveal narrative focus), and pragmatic tone (sentiment and subjectivity patterns across chapters). Together, these features act as cognitive ``fingerprints'' of expressions~\citep{biber1991variation, pennebaker2011secret}, reflecting how an author habitually selects words, constructs sentences, develops topics, and conveys emotional stance.

\noindent \textbf{Concept mappings}, grounded in the conceptual metaphor theory~\citep{lakoff2008metaphors}, reflect how individuals understand abstract domains through concrete source domains (e.g., \textsc{time is money}, or \textsc{love is a journey}). These mappings are cognitively salient because they capture deeper, often unconscious, structures of meaning-making. Recent studies provide a wealth of neural~\citep{mao2025eeg} and behavioral~\citep{mao2024unveiling} evidence to support the cognitive relevance of concept mappings. By modeling them, we aim to approximate how different authors structure their mental models of the world.

\noindent \textbf{Author profiles} include biographical and psychological dimensions such as persona, background, and personality traits (e.g., Big Five/OCEAN~\citep{goldberg1990alternative, john2008paradigm}). From a cognitive science perspective, lived experience and personality shape attentional focus, thematic preferences, and narrative voice~\citep{mcadams2006new}. While more indirect than linguistic or conceptual signals, profile-based features offer a higher-level view of the cognitive factors that potentially influence writing style.

Together, these representations span multiple layers of cognition, e.g., surface-level linguistic styles, mid-level conceptual systems, and high-level experiential profiles. 
By incorporating them in ICS, we provide a principled way to evaluate how well LLMs can approximate individualized cognitive patterns for storytelling.

\section{Dataset}

We curated a dataset of \textbf{five recently released novels} published in late 2024 and 2025, to reduce the chance that the evaluated LLMs had seen them during pretraining. Moreover, all selected novels were written by \textbf{well-known authors} (e.g., Stephen King, Suzanne Collins, Brandon Sanderson, \textit{et al.}). This choice ensures high-quality writing and allows us to gather rich metadata (e.g., biographies and personality insights) for building cognitive and stylistic profiles. To promote narrative diversity, we intentionally sampled across multiple genres and categories. Appendix~\ref{appendix:dataset} Table~\ref{tab:dataset-metadata} summarizes the book metadata (author, category, release date). For each novel, we used the opening chapters as \textbf{context data}, truncating them to match the smallest context window among the evaluated LLMs. The following chapters served as the \textbf{ground-truth continuations}, providing the human-written reference for comparison with model outputs. Appendix~\ref{appendix:dataset} Table~\ref{tab:dataset-statistics} shows text statistics, including chapter counts and text lengths for context and ground-truth segments.

\section{Methodology}
\begin{figure}[t]
    \centering
    \includegraphics[width=1\linewidth,,height=0.18\textheight]{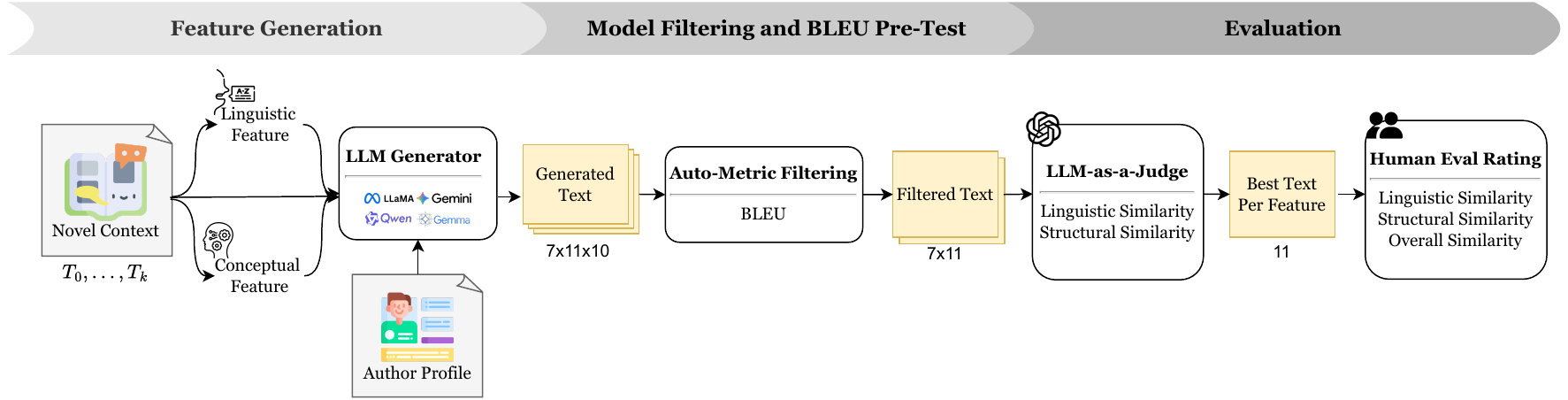}
    \caption{An overview of evaluation framework for Individualized Cognitive Simulation (ICS).}
    \label{fig:ics-framework}
\end{figure}

Our goal is to evaluate how different cognitive representations contribute to ICS for LLMs. To this end, we design a pipeline that benchmarks narrative continuations under multiple cognitive conditions. The framework, illustrated in Figure~\ref{fig:ics-framework}, consists of three stages: feature generation, model selection with BLEU pre-test, and evaluation. This design reduces human effort by using automatic metrics to pre-filter outputs, while targeted human judgment on the best candidates ensures evaluation remains both efficient and reliable.

\subsection{Feature Generation}

We designed eleven experimental conditions to systematically evaluate the utilities of different cognitive representations. As shown in Appendix~\ref{appendix:feature_condition}, these conditions are divided into two groups. The first group focuses on \textbf{single features}, including 1) a baseline with no added representation, 2) author persona, 3) author background, 4) author Big Five/OCEAN personality, 5) linguistic analysis of the preceding chapters, 6) concept mapping derived from the context, and 7) a combined author profile that merges persona, background, and personality traits. The second group introduces \textbf{multi-feature combinations}, where the three most cognitively grounded representations (author profile, linguistic features, and concept mappings) are combined in all possible pairwise and triplet configurations, yielding four additional setups.
These eleven conditions provide a controlled basis for comparing the influence of individual cognitive dimensions and their interactions. We describe each condition below in detail. Selected tools for feature generation are state-of-the-art. Sample prompt templates for all conditions are provided in Appendix~\ref{app:gen_prompts}.

\noindent \textbf{Baseline.} The baseline condition uses a simple continuation prompt that frames model as the novel's author and provides the opening chapters as input.

\noindent \textbf{Persona.} The persona condition captures anonymous demographic and identity attributes that may influence an author’s voice. These include age, nationality, gender, writing habits, and public-facing identity~\citep{song2019exploiting}. For example, a persona may describe an author as an older American novelist with a disciplined daily routine and outspoken social views, without disclosing the author’s actual name. These were common practices in LLM-based role-play tasks~\citep{xu2022cosplay,zheng2019personalizedtraits}. 

\noindent \textbf{Background.} The background condition captures an author’s academic experience and major life events that may shape their writing. This includes information such as education, professional training, formative jobs, and pivotal personal events (e.g., hardships, addictions, recoveries, or accidents). These elements provide higher-level cues about the themes that appear in an author’s work~\citep{baverstock2019what,lumen2021approaches}.

\noindent \textbf{Big Five/OCEAN personality.} The personality condition encodes the author’s psychological profile based on the Big Five (OCEAN) model~\citep{goldberg1990alternative, john2008paradigm}. Traits such as Openness to Experience and Conscientiousness provide cues about cognitive style, creativity, discipline, and narrative tendencies. In our setting, the author’s personality assessments are based on their public interviews, statements, and expert evaluations by two invited psychologists. The corresponding prompt conditions the model with this personality profile before continuing the story.

\noindent \textbf{Linguistic Features.} The linguistic condition provides the model with an explicit style guide derived from the opening chapters. Following the four dimensions introduced in Section~\ref{sec:preliminaries} (lexical, syntactic, semantic, pragmatic), we extract frequent vocabulary and bi-grams, part-of-speech distributions\citep{honnibal2020spacy} and sentence statistics, recurring topical words, and sentiment/subjectivity scores. These aggregated patterns are presented as a linguistic profile, which conditions the model to better match the author’s characteristic tone.

\noindent \textbf{Conceptual Features.} As introduced in Section~\ref{sec:preliminaries}, conceptual mappings capture how authors structure abstract ideas through concrete source domains. In our framework, we extract concept mappings from the context chapters using a computational metaphor processing system, MetaPro~\citep{mao2023metaproonline}. These concept mappings prompt the model to produce continuations that reflect the author’s conceptual system.

\noindent \textbf{Profile.} The author profile condition combines persona, background, and personality into a single prompt, offering a more holistic representation of the author’s biographical and psychological identity. These elements are combined because they are obtained from the authors' life experiences, in contrast to other features derived from the opening context.

\noindent \textbf{Multi-feature Combinations.} To test interactions between dimensions, we create prompts that combine author profile, linguistic features, and conceptual mappings. We include all pairwise combinations as well as one condition that integrates all three.  

In summary, each feature condition is implemented through a tailored prompt that implements cognitive representations in a consistent format. To maximize clarity and effectiveness, these prompts were refined with the assistance of state-of-the-art LLMs, ensuring natural and contextually appropriate instructions for every experimental condition.

\subsection{Model Filtering and BLEU Pre-Test}

All models in our pool were released before the earliest novel in our dataset, ensuring none of the texts could have been included in pretraining. To minimize this risk, we narrowed the selection to eight models across three major families—Google’s Gemma and Gemini, Meta’s Llama, and Alibaba’s Qwen—spanning multiple parameter scales for vertical comparison. Seven are open-source, while Gemini Pro 1.5 is a large proprietary system. To avoid evaluation bias, we excluded the GPT family, since GPT also serves as our evaluator. Since the smallest context window is 8,192 tokens, we standardized excerpt length and set the maximum output to $8192 - \text{input}$. A temperature of 0.8 was used to encourage creative writing. Appendix~\ref{appendix:model_info} Table~\ref{tab:models} summarizes model providers, sizes, context windows, release dates.

During model selection, we observed that LLMs are not robust across all tasks, and their exact capabilities for long-form narration remain uncertain. To ensure that the models included in our main experiments were able to produce valid outputs and mitigate prompt bias issues, we first conducted a BLEU-based filtering test. For each model, we tested 11 experimental conditions. In each setting, we generated 10 continuations for each of 5 novels. We then computed BLEU scores\citep{papineni2002bleu} against the ground-truth continuations and summarized per-model distributions (mean, std, min, max) in Appendix~\ref{appendix:bleu_pretest}. Since long-form narrative continuation admits many valid lexical phrasings, higher-order n\mbox{-}gram overlap with a single reference is rare; BLEU values are therefore expected to be near zero and we use BLEU solely as a coarse sanity filter (to flag empty, malformed, or off-language outputs) rather than as a quality metric. BLEU penalizes very short outputs, so BLEU\(=0\) flags truncated or unusable generations. On this scale, a mean BLEU of $0.0013$ is practically indistinguishable from $0$. Gemma~2B was the weakest (frequent empty/malformed outputs; BLEU$=0$) and was excluded; Appendix~\ref{app:gemma2b-example} also shows a well-formed Llama~3.2~3B continuation at near-zero BLEU ($0.0012$), so low BLEU need not imply low quality.

\subsection{Evaluation Dimensions}
For each model, we first selected the single highest BLEU-scoring response under each setting, yielding a total of $5 \text{ novels} \times 7 \text{ models} \times 11 \text{ settings} = 385$ candidate outputs. These responses were then assessed by an LLM-based evaluator. The evaluation dimensions included (1) linguistic style, (2) narrative structure, and (3) an overall rating score, computed as the average of the first two criteria. Next, we conducted a blinded human evaluation: for each setting, we took the top LLM-rated continuation per novel (yielding $5\times11=55$ items) and obtained 1--5 Likert ratings from five upper-division Literature/English majors on the same three dimensions with an independent overall score.

\subsubsection{linguistic style (LLM-based Automatic Rating)}
For linguistic style evaluation, we apply an LLM judge (\texttt{GPT-4 Turbo}) to 385 candidates, obtained by selecting the best-BLEU continuation for each model--setting--novel combination, yielding a single non-pathological representative and filtering degeneracies. The evaluator and linguistically informed prompt follow the work of~\citet{huang2024can-llms-identify-authorship}. The judge assesses whether a model’s continuation matches the author’s writing style, and assigns a 1--5 style-similarity score (5 = strongest same-author likelihood). The prompt explicitly instructs the judge to reason over surface-level stylistic cues, e.g., phrasal and modal verbs, punctuation, rare words and affixes, quantitative expressions, humor/sarcasm, and typographical patterns—rather than plot or entities (full prompt in Appendix~\ref{app:style-prompt}).

The evaluator returns a JSON object with a numeric score and a brief rationale. We take the numeric score as the linguistic-style score for that and aggregate across novels to obtain per-setting and per-model style results, to see which settings help on average and a model's overall style ability. Since BLEU measures lexical overlap rather than style, BLEU and the style score are complementary: the former is used only as a coarse sanity filter, while the latter targets authorial style directly.

\subsubsection{Narrative Structure (LLM-based Automatic Evaluation)}

Narrative structure in our framework is assessed through three components: \emph{event similarity, coverage, and ordering preservation}. Event similarity measures the quality of matched events, coverage tracks how many are aligned, and ordering captures sequence consistency.

\noindent \textbf{Event Representation and Similarity.} We decompose each paragraph into a sequence of events with \texttt{GPT-4 Turbo}. Each event \(e=(C,L,D)\) comprises a set of \textit{characters} \(C\), a \textit{location} string \(L\), and a short \textit{description} \(D\) of the main action. To compare generated and ground-truth events, we compute a weighted similarity that combines character overlap (Jaccard similarity after alias normalization), a location match based on token-level fuzzy matching, and semantic similarity between action descriptions (cosine similarity of sentence embeddings). The event-level score is a linear combination with fixed weights \((w_c,w_l,w_s)=(0.35,0.15,0.50)\). We prioritize semantic similarity of the main action (\(w_s\)) as most informative; character overlap (\(w_c\)) provides complementary evidence but can be noisy under aliasing; and location strings (\(w_l\)) are often sparse or under-specified: $S_{\text{event}} = w_c\, S_{\text{char}} \;+\; w_l\, S_{\text{loc}} \;+\; w_s\, S_{\text{sem}}$. This computation is detailed in Algorithm~\ref{alg:event-sim} (Appendix~\ref{appendix:structural}); the resulting values lie in \([0,1]\) and feed into downstream alignment and structural scoring. 


\noindent \textbf{Threshold-Based Alignment.} To compare sequences of events, we employ the Hungarian algorithm~\citep{kuhn1955hungarian,munkres1957algorithms} to align events between ground-truth and generated narratives (Algorithm~\ref{alg:threshold-hungarian}, Appendix~\ref{appendix:structural}). Unlike traditional applications that force a complete bipartite matching, we introduce a similarity threshold $\tau = 0.5$ and discard alignments below this threshold. This prevents spurious pairings of semantically unrelated events while preserving high-quality matches. The result is a set of aligned pairs $\mathcal{A}$ and an average event similarity $\bar{S}_{event}$ computed over these alignments.

\noindent \textbf{Structural Similarity Components.} From the filtered alignments, we compute three complementary measures (Algorithm~\ref{alg:struct-sim}, Appendix~\ref{appendix:structural}): (1) \textit{average event similarity}, reflecting the semantic quality of matched events, (2) \textit{coverage}, the proportion of events aligned across the two sequences, and (3) \textit{ordering preservation}, measured by Kendall’s $\tau$ correlation~\citep{kendall1945} between the relative positions of aligned events. Finally, we define the structural similarity score as a weighted combination: $S_{struct} = \alpha \,\bar{S}_{event} + \beta \,\text{Coverage} + \gamma \,\text{Ordering}$,
with default weights $\alpha = 0.6$, $\beta = 0.2$, and $\gamma = 0.2$ to reflect the relative reliability of each component: the semantic quality of matched events ($\bar{S}_{\text{event}}$) carries the greatest weight, while coverage and ordering provide complementary—but weaker—signals and are weighted equally. This formulation rewards continuations that match event content, maintain coverage, and preserve narrative order.


\subsubsection{Human Evaluation}


For each setting, we selected the response with the highest LLM-based \emph{overall} score, yielding one candidate per (novel, setting). We then conducted a blinded study with five Literature/English majors from top-ranked universities. Annotators compared each candidate with the ground-truth continuation on three dimensions: (1) \textit{linguistic style fidelity}, (2) \textit{narrative structure preservation}, and (3) an independent \textit{overall similarity} judgment, reflecting “same-author likelihood.”

Annotators viewed the ground-truth and model continuations side-by-side with randomized order; model identities and automatic scores were hidden. Instructions emphasized, for the style dimension, vocabulary/register, sentence structure and rhythm, tone/mood, and figurative or stylistic devices; and for the structure dimension, event coverage, character consistency, and event \emph{ordering}. Each dimension was scored on a 1--5 Likert scale with anchors (1~=\ very low similarity; 5~=\ very high similarity). The full survey, rubric, and examples appear in Appendix~\ref{appendix:human_eval}.


Quality control included randomization, attention checks on obvious mismatches, and exclusion of low-quality annotations. Final scores were averaged across five raters per item, then summarized as mean $\pm$ standard deviation across the five novels (55 items total).

\section{Results}

Table~\ref{tab:combined_llm_human} summarizes the results from both LLM-based automatic evaluation and human evaluation. In what follows, we first report the separate outcomes from each evaluation method, and then analyze why certain feature combinations—particularly Concept + Linguistic—performed best, while others such as Profile showed interesting discrepancies between LLM and human judgments.  

\noindent \textbf{LLM-based Automatic Evaluation Results.} As shown in Table~\ref{tab:combined_llm_human}, LLM-judge scores are reported for linguistic style (1--5) and structural similarity ([0,1]). The \emph{Concept + Linguistic} condition achieves the best overall rank$^{\dagger}$, driven by the top structural score \(0.167\,(0.257)\) and a strong linguistic rating \(3.057\,(1.608)\). Among single features, \emph{Profile} leads in linguistic style \(3.086\,(1.669)\) but lags in structure, with \emph{BigO Personality} showing the reverse pattern by ranking second in structure. Overall, combinations that integrate conceptual and linguistic signals outperform static profile-based cues. Structural scores are uniformly low (best \(0.167\,(0.257)\) on a [0,1] scale), but linguistic ratings are higher (roughly \(2.5\)–\(3.1\) out of 5), showing that models learn style more easily than structure.

\noindent \textbf{Human Evaluation Results.} As shown in Table~\ref{tab:combined_llm_human}, \emph{Concept + Linguistic} is best overall with the highest human scores for linguistic style \(3.40(1.19)\), structure \(2.60(0.80)\), and overall \(2.90\). Among single features, \emph{Concept} and \emph{Persona} tie for the best overall \(2.40\), while \emph{Profile} trails \(2.00\). Combinations that include \emph{Profile} underperform—\emph{Concept + Profile} \(1.90\), \emph{Profile + Linguistic} \(1.70\), and \emph{Profile + Concept + Linguistic} lowest at \(1.40\). Across all settings, structure is consistently harder than style (structure \(\approx 1.20\)–\(2.60\) vs.\ style \(\approx 1.60\)–\(3.40\)), reinforcing that human raters perceive stronger stylistic than structural alignment. The observed standard deviations (style \(\sim\)0.8–1.4; structure \(\sim\)0.3–1.0) indicate substantial between-novel variability, but the overall pattern mirrors the LLM-based results: integrating conceptual cues with linguistic features yields the most human-preferred continuations, while profile-heavy prompts do not.

\noindent \textbf{Why Concept + Linguistic Performs Best?} The superior performance of the \emph{Concept + Linguistic} combination can be understood from the perspective of feature complementarity. As discussed in the preliminary analysis, Linguistic features capture surface-level style and syntax–semantics, while concept mapping reflects mid-level conceptual and pragmatic structures. By integrating these two dimensions, the model benefits from both coherent linguistic expression on the surface and deeper conceptual grounding beneath. In contrast, Profile-based combinations do not achieve similar improvements. Profile information, while sometimes useful on its own for capturing an author’s background or thought patterns, provides only indirect signals for text generation. When combined with linguistic or conceptual features, it can dilute and may even introduce noisy or inconsistent information. Thus, unlike profile-based combinations, the Concept + Linguistic pairing offers clear and synergistic guidance, yielding the best results. Sample ground truth and generation results are shown in Appendix~\ref{appendix:sample_results}.

\noindent \textbf{Why the Profile Feature Ranks Higher Linguistically in LLMs?} Interestingly, the Profile feature received relatively high ratings from the LLM in the linguistic dimension, but its human evaluation scores were much less favorable. One possible explanation is that the LLM-as-judge framework may have blind spots in linguistic assessment, particularly for content conditioned on profile information. While humans judged Profile to be weaker than Concept, Linguistic, or Persona, the LLM tended to score it more generously, suggesting a mismatch between automated and human judgments in this dimension. This discrepancy may also help explain why Profile performs poorly in combination with other features: because the profile signal is less directly tied to surface-level stylistic or conceptual cues, adding it can introduce noise rather than enhance linguistic quality.

\begin{table*}[t]
\centering
\scriptsize
\setlength{\tabcolsep}{4pt}
\caption{Combined LLM-based and human evaluations. Values are mean(std). Overall ranks: LLM = average of linguistic \& structural ranks; Human = rank by human overall. $^{\dagger}$Best overall.}
\label{tab:combined_llm_human}
\begin{tabular}{lccccccc}
\toprule
\multirow{2}{*}{\textbf{Setting}} &
\multicolumn{2}{c}{\textbf{LLM}} &
\multicolumn{3}{c}{\textbf{Human (1--5)}} &
\multicolumn{2}{c}{\textbf{Overall Rank}} \\
\cmidrule(lr){2-3}\cmidrule(lr){4-6}\cmidrule(lr){7-8}
& \textbf{Linguistic} & \textbf{Structure} & \textbf{Linguistic} & \textbf{Structure} & \textbf{Overall} & \textbf{LLM} & \textbf{Human} \\
\midrule
\multicolumn{8}{c}{\textit{Single Features}} \\
\midrule
Profile                  & \textbf{3.09(1.67)} & 0.107(0.22) & 2.40(1.36) & 1.70(0.81) & 2.00 & 2 & 5 \\
Background               & 2.94(1.59) & 0.055(0.16) & 2.00(1.26) & 1.40(0.66) & 1.60 & 7 & 10 \\
Concept                  & 2.94(1.59) & 0.103(0.21) & 2.80(1.10) & 1.60(0.98) & 2.40 & 5 & 2 \\
Base                     & 2.91(1.58) & 0.098(0.19) & 2.40(1.10) & 1.40(0.49) & 1.70 & 6 & 7 \\
Linguistic               & 2.83(1.76) & 0.113(0.22) & 2.60(1.19) & 1.80(1.00) & 2.20 & 3 & 4 \\
BigO Personality         & 2.71(1.34) & \underline{0.126(0.25)} & 2.00(0.83) & 1.70(0.65) & 1.70 & 3 & 7 \\
Persona                  & 2.66(1.31) & 0.112(0.23) & 2.90(1.11) & 1.80(0.78) & 2.40 & 7 & 2 \\
\midrule
\multicolumn{8}{c}{\textit{Multi-feature Combinations}} \\
\midrule
\textbf{Concept + Linguistic}$^{\dagger}$ & \underline{3.06(1.61)} & \textbf{0.167(0.26)} & \textbf{3.40(1.19)} & \textbf{2.60(0.80)} & \textbf{2.90} & \textbf{1} & \textbf{1} \\
Concept + Profile         & 2.83(1.52) & 0.084(0.17) & 2.60(0.99) & 1.20(0.50) & 1.90 & 10 & 6 \\
Profile + Linguistic      & 2.66(1.55) & 0.090(0.21) & 2.40(1.00) & 1.20(0.30) & 1.70 & 11 & 7 \\
Profile + Concept + Linguistic & 2.51(1.54) & 0.122(0.23) & 1.60(1.02) & 1.20(0.40) & 1.40 & 7 & 11 \\
\bottomrule
\end{tabular}
\end{table*}

\subsection{Linguistic Style Analysis}

\begin{figure}[t]
    \centering
    \includegraphics[width=1\linewidth]{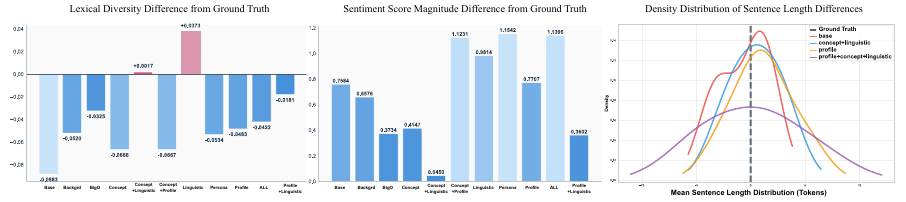}
    \caption{Linguistic style analysis of lexical diversity, sentiment, and sentence length. Concept + Linguistic achieves the closest match to ground truth.}
    \label{fig:linguistic}
\end{figure}

We further analyze the generated texts in terms of surface-level linguistic properties, focusing on lexical diversity, sentiment, and sentence length (Figure~\ref{fig:linguistic}).  

For lexical diversity, measured by normalized type–token ratio (TTR), most settings fall below the ground truth, reflecting reduced vocabulary variety. The \emph{Concept + Linguistic} combination achieves the closest match to human texts (\(+0.0017\) difference), while the Linguistic feature alone yields the highest diversity (\(+0.0373\)). It shows that although purely linguistic prompts promote greater lexical variety, the Concept + Linguistic combination strikes a better balance by aligning most closely with the human-authored baseline.  

Sentiment analysis, conducted using SenticNet~\citep{camnt8}, shows a similar pattern. While lexical diversity relies on type–token ratios, sentiment was measured with SenticNet polarity scores, using the absolute difference from the ground truth. Most systems differ noticeably from the ground truth, showing larger shifts in sentiment intensity regardless of direction. The \emph{Concept + Linguistic} combination again produces the smallest difference \(+0.0450\), showing that combining conceptual and linguistic features helps the model keep sentiment closer to the ground truth than any other feature set. In contrast, Profile-based combinations exhibit much larger deviations, indicating noisier sentiment control.

We analyzed sentence lengths by comparing their distributions to the ground truth. The peak of the \emph{Concept + Linguistic} curve is closest to the human distribution, showing that this combination best reproduces the natural rhythm of human writing. Both the \emph{Concept + Linguistic} and \emph{Persona} curves are narrower than other settings, suggesting a more concentrated distribution around typical sentence lengths. By contrast, the curve for the all-feature combination diverges much more strongly, indicating that its generations differ substantially from the ground truth in sentence-level structure.

\subsection{Narration Structure Analysis}

\begin{figure}[t]
    \centering
    \includegraphics[width=0.7\linewidth]{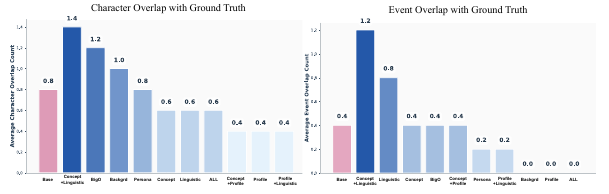}
    \caption{Structural analysis (character and event overlap with ground truth). Concept + Linguistic yields the strongest alignment.}
    \label{fig:structure}
\end{figure}

Character overlap was measured by first using \texttt{GPT-4 Turbo} to extract character mentions from both the generated texts and the ground truth. We then collected all aliases of each character appearing in the ground truth and counted the overlapping characters present in each generation. Each novel contains about four characters on average, and the reported overlap scores represent averages across all novels. As shown in Figure~\ref{fig:structure} (left), most features yield relatively low overlap, with values below the baseline score of 0.8. The \emph{Concept + Linguistic} combination stands out, achieving the highest overlap (1.4), followed by BigO (1.2) and Background (1.0). In contrast, profile-based and all-feature combinations perform the worst, with overlap values as low as 0.4, suggesting that these settings introduce noise and reduce structural alignment.

Event overlap was determined by first calculating the semantic similarity between generated texts and the ground truth using BERTscore, followed by human justification to verify overlapping events. Each novel contains about five events on average, and the reported overlap scores represent averages across all novels. As shown in Figure~\ref{fig:structure}(right), the \emph{Concept + Linguistic} combination achieves the highest overlap (1.2), followed by BigO (1.0) and Background (0.9). Most other settings remain below the baseline of 0.8, with profile-based and all-feature combinations again performing the worst (as low as 0.3). These results suggest that while event-level alignment remains more challenging than character overlap, conceptual and linguistic cues together help the model more reliably reuse key events from the ground truth.

\subsection{Model Performance Analysis}
Table~\ref{tab:combined_models} summarizes model performance across the two LLM-based evaluation dimensions. Overall, \emph{Gemini Pro 1.5} performs best, achieving the top linguistic ranking (3.46) and the second-best structural similarity score (0.123). In contrast, \emph{Gemma-7B} is the weakest performer, ranking last in linguistic style (2.11) and near the bottom in structure (0.096). Interestingly, \emph{Llama-3.2 3B Instruct} stands out by obtaining the highest structural similarity (0.144), despite being only mid-ranked on linguistic style. Examining scale effects within model families, larger models tend to perform better on surface-level linguistic style but not necessarily on deeper structural similarity. For example, in the Qwen family, the larger \emph{7B} model achieves stronger linguistic fluency (3.42 vs. 2.18) while the smaller \emph{1.8B} model shows better structural alignment (0.117 vs. 0.068). These results indicate that scaling benefits style more than structure, with structural similarity relying more on training and alignment.

\begin{table}[t]
\centering
\scriptsize
\caption{Comparison of model performance across two LLM-based evaluation dimensions. Linguistic style scores are on a 1--5 scale, while structural similarity scores are in [0,1]. Rankings are global across all models.}
\label{tab:combined_models}
\begin{tabular}{lcccccc}
\toprule
\textbf{Model} & \textbf{Ling. Rank} & \textbf{Mean (Std)} & \textbf{Struct. Rank} & \textbf{Mean (Std)} \\
\midrule
Gemini Pro 1.5         & \textbf{1} & \textbf{3.46} (1.57) & 2 & 0.123 (0.231) \\
Qwen-1.5 7B            & 2 & 3.42 (1.51) & 7 & 0.068 (0.164) \\
Llama-3.2 3B Instruct  & 3 & 3.18 (1.47) & \textbf{1} & \textbf{0.144} (0.241) \\
Llama-3.2 1B Instruct & 4 & 3.00 (1.44) & 6 & 0.094 (0.196) \\
Qwen-1.5 4B            & 5 & 2.62 (1.50) & 4 & 0.101 (0.221) \\
Qwen-1.5 1.8B          & 6 & 2.18 (1.29) & 3 & 0.117 (0.232) \\
Gemma-7B              & 7 & 2.11 (1.38) & 5 & 0.096 (0.211) \\
\bottomrule
\end{tabular}
\end{table}

\section{Conclusion}

This work introduces ICS, where models approximate an author’s cognitive and stylistic processes in narrative continuation. We proposed an evaluation framework spanning 11 conditions, integrating linguistic, conceptual, and profile-based signals, assessed with both LLM-based metrics and human judgments. Results show that combining concept mappings with linguistic features is most effective, while LLMs remain limited to surface-level style and struggle with deeper cognitive traits. 

These findings highlight the need for new training, prompting, and decoding methods to advance beyond stylistic mimicry toward more faithful cognitive simulation.
Models capture surface-level linguistic style but struggle with deeper cognitive traits, and scaling alone does not close this gap. Larger models improve stylistic fluency but not cognitive fidelity, suggesting that progress in ICS will require new training strategies and data design rather than size increases alone.




\section*{Ethics Statement}

This research focuses on individualized cognitive simulation (ICS) using large language models. All data used in our experiments are publicly available, consisting of published literary works and publicly accessible author information (e.g., Wikipedia entries and public interviews). No private or proprietary data were used. Our experiments do not involve human subjects, sensitive personal information, or interventions that may pose harm. Human evaluations were conducted by recruited annotators under informed consent, with no demographic or identifying information collected. 

\bibliography{iclr2026_conference}
\bibliographystyle{iclr2026_conference}

\newpage
\appendix

\section{Dataset Information}
\label{appendix:dataset}

\begin{table}[H]
\centering
\small
\begin{threeparttable}
\caption{Dataset overview - Book metadata.}
\label{tab:dataset-metadata}
\begin{tabular}{L{4.2cm} L{3.5cm} L{3.5cm} C{2.5cm}}
\toprule
\textbf{Book Title} & \textbf{Author} & \textbf{Category} & \textbf{Release Date} \\
\midrule
Sunrise on the Reaping & Suzanne Collins & Adventure\&Fantasy & 03/08/2025 \\
Never Flinch & Stephen King & Horror & 12/06/2024 \\
Wind and Truth & Brandon Sanderson & Fantasy\&Sci-fi & 12/06/2024 \\
What Does It Feel Like? & Sophie Kinsella & Autobiographical & 10/04/2024 \\
The Mighty Red & Louise Erdrich & Literary & 10/01/2024 \\
\bottomrule
\end{tabular}
\begin{tablenotes}[flushleft]
\footnotesize
\item \textbf{Note:} Dates are in mm/dd/yyyy format.
\end{tablenotes}
\end{threeparttable}
\end{table}

\begin{table}[H]
\centering
\small
\begin{threeparttable}
\caption{Dataset overview—Text statistics.}
\label{tab:dataset-statistics}
\begin{tabular}{L{3.8cm} C{2.7cm} C{2.7cm} C{2.7cm} C{2.7cm}}
\toprule
\textbf{Book Title} &  \textbf{Context Chapters} & \textbf{Context (W/T)} & \textbf{Ground Truth Chapters} & \textbf{Ground Truth (W/T)} \\
\midrule
Sunrise on the Reaping   & Ch.~1 (first half) & 4{,}633 / 5{,}937 & Ch.~1 (second half) & 2{,}267 / 2{,}929 \\
Never Flinch             & Ch.~1--3           & 4{,}688 / 6{,}418 & Ch.~4               & 1{,}122 / 1{,}322 \\
Wind and Truth           & Ch.~1              & 4{,}874 / 6{,}675 & Ch.~2               & 2{,}573 / 3{,}529 \\
What Does It Feel Like?   & Ch.~1--3           & 4{,}959 / 6{,}896 & Ch.~4               & 1{,}127 / 1{,}482 \\
The Mighty Red          & Ch.~1--4           & 5{,}076 / 6{,}974 & Ch.~5               & 2{,}297 / 3{,}231 \\
\bottomrule
\end{tabular}
\begin{tablenotes}[flushleft]
\footnotesize
\item \textbf{Abbreviations:} W/T = words/tokens. “Context” is the input to the LLMs.
\end{tablenotes}
\end{threeparttable}
\end{table}

\section{Overview of Feature Conditions}
\label{appendix:feature_condition}

\begin{figure}[H]
    \centering
    \includegraphics[width=0.5\linewidth]{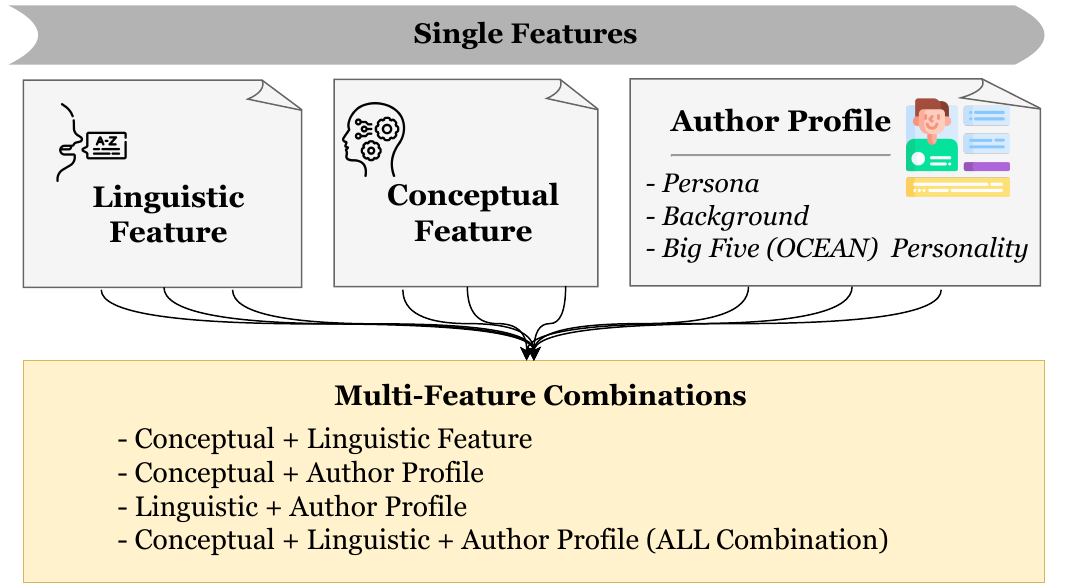}
    \caption{Overview of feature conditions used for individualized cognitive simulation (ICS)}
    \label{fig:features}
\end{figure}

\section{Generation Prompts}
\label{app:gen_prompts}

\begin{tcolorbox}[
  enhanced, breakable, colback=gray!5, colframe=black!50, sharp corners,
  boxrule=0.5pt, left=6pt, right=6pt, top=6pt, bottom=6pt,
  width=\linewidth,
  before skip=2pt, after skip=2pt,
  enlarge left by=0mm,
  enlarge right by=0mm
]
\textbf{Baseline Prompt}\\
\small\ttfamily
\\
You are the novel author of `\{novel\_name\}'. \\
You are given the opening chapters of this novel. \\
Your job is to continue writing and only output the narration. \\
The narration should be in the style of the novel. \\

\{novel\_context\_data\}
\end{tcolorbox}

$ $

\begin{tcolorbox}[
  enhanced, breakable, 
  colback=gray!5, colframe=black!50, sharp corners,
  boxrule=0.5pt, left=6pt, right=6pt, top=6pt, bottom=6pt,
  before skip=2pt, after skip=2pt,
  width=\linewidth,      
  enlarge left by=0mm,
  enlarge right by=0mm
]
\textbf{Persona Sample Prompt}\\
\small\ttfamily
\\
You are the author of `Never Flinch'. \\
Here is your persona profile: \\
The author is a 77-year-old American male novelist whose disciplined 
"write-every-day" habit—he still aims for roughly a thousand words daily— 
reflects a methodical, workmanlike personality beneath his sardonic Maine wit. \\
Publicly, he balances introverted creativity with an outspoken progressive streak 
on X/Twitter and in interviews, railing against censorship and climate-change denial. \\

You are given the opening chapters of this novel. \\
Your job is to continue writing and only output the narration. \\
The narration should be in the style of the novel. \\

\{never\_flinch\_context\_data\}\\

\textit{(The persona information was collected from author's Wikipedia entry.)}
\end{tcolorbox}

$ $

\begin{tcolorbox}[
  enhanced, breakable, colback=gray!5, colframe=black!50, sharp corners,
  boxrule=0.5pt, left=6pt, right=6pt, top=6pt, bottom=6pt,
  width=\linewidth, 
  before skip=2pt, after skip=2pt,
  enlarge left by=0mm,   
  enlarge right by=0mm,
  before skip=1pt, after skip=1pt,
]
\textbf{Background Sample Prompt} \\
\small\ttfamily
\\
You are the author of `Never Flinch'. \\
Here is your background: \\
Raised in deep-rooted New England poverty after his father abandoned the family,
the author (born Portland, Maine, 1947) cycled through a dozen states before
settling in Durham, ME, where he devoured pulp, wrote prodigiously and, at the
University of Maine, earned a B.A. in English (1970) and met fellow writer, whom he married the next year. \\
To keep food on the table he taught high-school English, cleaned girls’ locker-room
showers and toiled in an industrial laundry—gritty day jobs that later sparked
\emph{Carrie} and stories like “The Mangler.” \\
... \\
Today, after 60-plus books and an estimated 400 million copies sold, the author’s
life story—poverty, blue-collar grit, addiction, resilience—continues to feed the
nightmares and moral undercurrents that define his fiction. \\

You are given the opening chapters of this novel. \\
Your job is to continue writing and only output the narration. \\
The narration should be in the style of the novel. \\

\{never\_flinch\_context\_data\}\\

\textit{(The background information was collected from author's Wikipedia entry and publicly available interviews.)}
\end{tcolorbox}

$ $

\begin{tcolorbox}[
  enhanced, breakable, colback=gray!5, colframe=black!50, sharp corners,
  boxrule=0.5pt, left=6pt, right=6pt, top=6pt, bottom=6pt,
  width=\linewidth,
  before skip=2pt, 
  after skip=2pt,
  enlarge left by=0mm,
  enlarge right by=0mm
]
\textbf{Personality Sample Prompt}\\
\small\ttfamily
\\
You are the author of `Never Flinch'. \\
Here is your Big Five/OCEAN personality profile (all traits are high): \\
Openness to Experience: The extent to which someone is curious, imaginative, 
and open to new ideas and experiences. \\
Conscientiousness: The degree of self-discipline, organization, 
and goal-directed behavior. \\

You are given the opening chapters of this novel. \\
Your job is to continue writing and only output the narration. \\
The narration should be in the style of the novel. \\

\{never\_flinch\_data\}\\

\textit{(Author’s personality information was collected by ChatGPT-5, after providing relevant interviews of the author.)}
\end{tcolorbox}
\begin{tcolorbox}[
  enhanced, breakable, colback=gray!5, colframe=black!50, sharp corners,
  boxrule=0.5pt, left=6pt, right=6pt, top=6pt, bottom=6pt,
  width=\linewidth,
  before skip=2pt, after skip=2pt,
  enlarge left by=0mm,
  enlarge right by=0mm
]
\textbf{Linguistic Sample Prompt}\\
\small\ttfamily
\\
You are the author of `Never Flinch'. \\
You are given the opening chapters of this novel, followed by a stylistic analysis. \\
Your job is to continue writing and only output the narration. \\
The continuation should strictly match the established voice, tone, pacing, 
and style of the original text. \\

PREVIOUS CHAPTERS: \\
\{never\_flinch\_conetxt\_data\} \\

LINGUISTIC STYLE GUIDE: \\

Lexical Style: \\
Frequent vocabulary: holly, izzy, says, duffrey, like, know, barbara, got, think, say, 
tolliver, time, going, innocent, lewis, said, wilson, big.\\
Common bi-grams: holly says, izzy says, cary tolliver, alan duffrey, blackstone rule, 
sista bessie, ace detective, buckeye brandon. \\

Syntactic Style: \\
High noun and verb density for concrete and action-driven sentences. \\
Frequent pronouns maintain close narrative perspective. \\
Named entities recur (characters, places, objects). \\
Adjectives/adverbs are minimal and efficient. \\
Short sentences with clear punctuation define rhythm. \\
Frequent use of determiners and prepositions. \\
Auxiliary verbs and conjunctions support tense and complexity. \\
Occasional interjections add emotional realism. \\

Semantic Themes: \\
Recurring topic words across chapters: \\
Chapter 1: izzy, duffrey, innocent, tolliver, warwick \\
Chapter 2: holly, duffrey, tacos, tolliver \\
Chapter 3: barbara, bessie, gospel, sista \\
Chapter 4: nurse, flash, solarium, cary \\

Pragmatic Tone: \\
Chapter 1 – Polarity: 0.012, Subjectivity: 0.202 \\
Chapter 2 – Polarity: 0.037, Subjectivity: 0.197 \\
Chapter 3 – Polarity: 0.065, Subjectivity: 0.268 \\
Chapter 4 – Polarity: 0.001, Subjectivity: 0.242 \\

INSTRUCTION: \\
Write the continuation of the story. \\
Maintain consistency in character voice, sentence rhythm, lexical choices, 
and overall tone. \\
Output only the next chapter of the narrative and DO NOT output the chapter number. \\
\end{tcolorbox}%

$ $

\begin{tcolorbox}[
  enhanced, breakable, colback=gray!5, colframe=black!50, sharp corners,
  boxrule=0.5pt, left=6pt, right=6pt, top=6pt, bottom=6pt,
  width=\linewidth,
  before skip=2pt, after skip=2pt,
  enlarge left by=0mm,
  enlarge right by=0mm
]
\textbf{Conceptual Mappings Sample Prompt}\\
\small\ttfamily
\\
You are the author of 'Never Flinch'. \\

The following concept mappings pairs are provided as thematic inspiration: \\
\#\#\# Concept Mappings Pairs\#\#\# \\
`POSSESSION' is `ACTION' \\
`CRUELTY' is `FEELING' \\
`SIZE' is `QUALITY' \\
`TEMPORAL\_PROPERTY' is `FORCE' \\
`WAKEFULNESS' is `UPRIGHTNESS' \\
`MOVEMENT' is `STATE' \\
...\\
\#\#\# End of Concept Mappings \#\#\# \\

Below is the beginning of your novel: \\
\#\#\# Opening Chapters \#\#\# \\
\{never\_flinch\_Context\_data\} \\
\#\#\# End of Opening Chapters \#\#\# \\

Your task is to continue writing the narration in the same tone and style. \\
Incorporate the above concept mappings where thematically appropriate. \\
Do not explain or label anything — only output the next part of the narration. \\
Do not include chapter number. \\

\#\#\# Continue the Narration Below \#\#\#
\end{tcolorbox}%

$ $

\begin{tcolorbox}[
  enhanced, breakable, colback=gray!5, colframe=black!50, sharp corners,
  boxrule=0.5pt, left=6pt, right=6pt, top=6pt, bottom=6pt,
  width=\linewidth,
  before skip=2pt, 
  after skip=2pt,
  enlarge left by=0mm,
  enlarge right by=0mm
]
\textbf{Author Profile Prompt}\\
\small\ttfamily
\\
You are the author of `\{novel\_name\}'. \\

Here is your persona profile: \\
`\{Author\_Persona\}' \\

Here is your background: \\
`\{Author\_Background\}' \\

Here is your Big Five/OCEAN personality profile (all traits are high): \\
`\{Author\_Personality\}' \\

You are given the opening chapters of this novel. \\
Your job is to continue writing and only output the narration. \\
The narration should be in the style of the novel. \\

\{novel\_context\_data\}
\end{tcolorbox}%

$ $

\begin{tcolorbox}[
  enhanced, breakable, colback=gray!5, colframe=black!50, sharp corners,
  boxrule=0.5pt, left=6pt, right=6pt, top=6pt, bottom=6pt,
  width=\linewidth,
  before skip=2pt, 
  after skip=2pt,
  enlarge left by=0mm,
  enlarge right by=0mm
]
\textbf{All-Combination Prompt}\\
\small\ttfamily
\\
You are the author of `\{novel\_name\}'. \\

Here is your persona profile: \\
`\{Author\_Persona\}' \\

Here is your background: \\
`\{Author\_Background\}' \\

Here is your Big Five/OCEAN personality profile (all traits are high): \\
`\{Author\_Personality\}' \\

Here is the concept mappings pairs: \\
`\{Concept\_Mappings\_Pairs\}' \\

Here is the linguistic style guide: \\
`\{Linguistic\_Features\}' \\

You are given the opening chapters of this novel. \\
Your job is to continue writing and only output the narration. \\
The narration should be in the style of the novel. \\

\{novel\_context\_data\}
\end{tcolorbox}

\section{Model Information}
\label{appendix:model_info}
\begin{table}[H]
\centering
\small
\begin{threeparttable}
\caption{Models used in our experiments.}
\label{tab:models}
\begin{tabular}{L{3.5cm} C{2.3cm} C{3cm} C{3cm} C{2.6cm}}
\toprule
\textbf{Model Name} & \textbf{Provider} & \textbf{Context Window} & \textbf{Parameters} & \textbf{Release Date} \\
\midrule
\model{Gemma-2b}               & Google  & 8{,}192   & 2B      & 2024-02-21 \\
\model{Gemma-7b}               & Google  & 8{,}192   & 7B      & 2024-02-21 \\
\model{Gemini-pro-1.5}         & Google  & 2{,}000k  & Unknown\textsuperscript{\dag} & 2024-04-09 \\
\model{Llama-3.2-1B-Instruct}  & Meta    & 128k      & 1B      & 2023-12-01 \\
\model{Llama-3.2-3B-Instruct}  & Meta    & 128k      & 3B      & 2024-09-25 \\
\model{Qwen1.5-1.8B}           & Alibaba & 32k       & 1.8B    & 2024-02-01 \\
\model{Qwen1.5-4B}             & Alibaba & 32k       & 4B      & 2024-02-01 \\
\model{Qwen1.5-7B}             & Alibaba & 32k       & 7B      & 2024-02-01 \\
\bottomrule
\end{tabular}
\begin{tablenotes}[flushleft]
\footnotesize
\item \textsuperscript{\dag}\,Official parameter count not publicly released; widely understood to be very large.
\end{tablenotes}
\end{threeparttable}
\end{table}

\section{BLEU Score Statistic from the Pre-Test}
\label{appendix:bleu_pretest}
\begin{table}[H]
\centering
\small
\caption{BLEU score statistics from the pre-test.
Gemma 7B and Gemma 2B produced only 540/550 responses due to empty outputs in the concept mapping condition.}
\begin{tabular}{L{3.8cm} c c c c c}
\toprule
\textbf{Model} & \textbf{Mean} & \textbf{Std} & \textbf{Max} & \textbf{Min} \\
\midrule
\model{Qwen1.5-4B}              & 0.0013 & 0.0005 & 0.0067 & 0.0000 \\
\model{Qwen1.5-7B}              & 0.0013 & 0.0006 & 0.0072 & 0.0000 \\
\model{Qwen1.5-1.8B}            & 0.0011 & 0.0009 & 0.0080 & 0.0000 \\
\model{Gemma-7B}                & 0.0008 & 0.0007 & 0.0051 & 0.0000 \\
\model{Llama-3.2-3B-Instruct}   & 0.0007 & 0.0004 & 0.0027 & 0.0000 \\
\model{Llama-3.2-1B-Instruct}   & 0.0005 & 0.0003 & 0.0037 & 0.0000 \\
\model{Gemini-Pro-1.5}          & 0.0001 & 0.0001 & 0.0009 & 0.0000 \\
\rowcolor{gray!15} \model{Gemma-2B (removed)} & 0.0000 & 0.0001 & 0.0007 & 0.0000 \\
\bottomrule
\end{tabular}
\end{table}



\lstset{
  basicstyle=\ttfamily\small,
  frame=single,
  breaklines=True,
  breakatwhitespace=true,
  breakautoindent=false,
  breakindent=0pt,
  columns=fullflexible,
  keepspaces=true,
  upquote=true,
  postbreak=\mbox{},  
}

\section{Example Outputs: Malformed (Gemma 2B) vs.\ Usable (Llama 3.2 3B)}
\label{app:gemma2b-example}

\noindent Gemma 2B --- BLEU = 0 (malformed):
\begin{mdframed}
\noindent\texttt{<h3><h3>>
] ]: the the the the the the, s t- ']
' ' 'm:body the a a a n of the first ]ES']}
\end{mdframed}

\noindent Llama 3.2 3B --- BLEU = 0.0012 (usable):
\begin{mdframed}
\noindent\texttt{They enter the solarium, a glass-enclosed room filled with plants and sunlight pouring in. Cary Tolliver lies on a bench, wearing a pair of reading glasses perched on the end of his nose.
He's got a book open in front of him and his eyes are half-closed, lost in the story. He looks up, and his eyes snap open. "Oh, detectives. What a surprise. I was hoping you'd come." Izzy and Tom exchange a look. "We didn't know you were expecting us," Izzy says. "I was hoping you'd show up. Come sit down." He gestures to the bench beside him. Tom looks around the room. "Is there anything you want us to..." "Just the truth, Cary. That's all I'm asking for." Izzy nods. "We know about the confession letter, and we know about Buckeye Brandon." He nods, eyes still on her. "I wrote that letter. Yes. I was hoping it would make a difference, but I guess it wasn't enough." "What do you want to tell us now, Cary? You said you came forward after you got your diagnosis...
[truncated for space]}
\end{mdframed}

\section{Linguistic Style Evaluation Prompt}
\label{app:style-prompt}

\begin{lstlisting}[basicstyle=\ttfamily\small, frame=single, rulecolor=\color{lstframe},
  breaklines=true, columns=fullflexible, keepspaces=true,backgroundcolor=\color{lstbg}]
Verify whether two input texts were written by the same author.
Analyze writing style only; ignore topic and content.
Base your reasoning on linguistic features such as phrasal verbs, modal verbs, punctuation,
rare words, affixes, quantities, humor, sarcasm, typographical patterns, and misspellings.

Respond on a 1-5 scale (5 = highly likely same author; 1 = highly unlikely).
Output JSON only, with keys "score" (number) and "rationale" (string).

Input text 1: <ORIGINAL_TEXT>
Input text 2: <MODEL_RESPONSE_TEXT>
\end{lstlisting}


\section{Structural Similarity Algorithms}
\label{appendix:structural}

\begingroup
\setlength{\textfloatsep}{6pt}

\begin{algorithm}[H]
\caption{EventSimilarity with Aliases}
\label{alg:event-sim}
\begin{algorithmic}[1]
\Require Ground-truth event $e_g$, generated event $e_h$, alias map $\mathcal{M}$, sentence encoder $\mathrm{Emb}(\cdot)$; weights $(w_c,w_l,w_s)=(0.35,0.15,0.50)$; fuzzy threshold $\tau_{\text{loc}}=0.8$; coarse match value $c=0.5$
\Ensure Event similarity $S_{\text{event}}\in[0,1]$
\State $C_g \gets \{\textsc{Canon}(x;\mathcal{M}) : x \in e_g.\text{characters}\}$; $C_h \gets \{\textsc{Canon}(x;\mathcal{M}) : x \in e_h.\text{characters}\}$
\State $S_{\text{char}} \gets \dfrac{|C_g \cap C_h|}{|C_g \cup C_h|}$ \Comment{Jaccard similarity}
\State $L_g \gets \text{trim}(e_g.\text{location})$; $L_h \gets \text{trim}(e_h.\text{location})$
\State $s \gets \textsc{DiceTokens}(L_g,L_h)$ \Comment{Sørensen–Dice over token sets}
\State $S_{\text{loc}} \gets \begin{cases}
1, & \text{if } \textsc{Lower}(L_g)=\textsc{Lower}(L_h) \\
c, & \text{if } s \ge \tau_{\text{loc}} \\
0, & \text{otherwise}
\end{cases}$
\State $d_g \gets e_g.\text{description or main\_event}$; $d_h \gets e_h.\text{description or main\_event}$
\State $u \gets \mathrm{Emb}(d_g)$; $v \gets \mathrm{Emb}(d_h)$; \quad $S_{\text{sem}} \gets \dfrac{u\cdot v}{\|u\|\|v\|}$ \Comment{cosine similarity}
\State \Return $S_{\text{event}} \gets w_c S_{\text{char}} + w_l S_{\text{loc}} + w_s S_{\text{sem}}$
\end{algorithmic}
\end{algorithm}

\begin{algorithm}[H]
\caption{Thresholded Hungarian Alignment}
\label{alg:threshold-hungarian}
\begin{algorithmic}[1]
\Require GT events $G=\{g_i\}_{i=1}^{n_g}$, GEN events $H=\{h_j\}_{j=1}^{n_h}$, novel name $z$, threshold $\tau=0.5$, encoder $\mathrm{Emb}(\cdot)$
\Ensure Aligned pairs $\mathcal{A}$, average event similarity $\bar{S}_{\text{event}}$
\State $\mathcal{M} \gets \textsc{BuildAliasMap}(z)$
\State Build similarity matrix $S\in\mathbb{R}^{n_g\times n_h}$ with $S_{ij}\gets \textsc{EventSimilarity}(g_i,h_j,\mathcal{M},\mathrm{Emb})$ \Comment{Alg.~\ref{alg:event-sim}}
\State $C \gets -S$ \Comment{cost matrix for minimization}
\State $(\mathbf{r},\mathbf{c}) \gets \textsc{Hungarian}(C)$ \Comment{linear\_sum\_assignment}
\State $\mathcal{A} \gets \{(r_k,c_k)\,:\, S_{r_k c_k} \ge \tau\}$
\State $\bar{S}_{\text{event}} \gets \begin{cases}
\dfrac{1}{|\mathcal{A}|}\sum\limits_{(i,j)\in\mathcal{A}} S_{ij}, & |\mathcal{A}|>0\\[6pt]
0, & \text{otherwise}
\end{cases}$
\State \Return $\mathcal{A}, \bar{S}_{\text{event}}$
\end{algorithmic}
\end{algorithm}

\begin{algorithm}[H]
\caption{Structural Similarity}
\label{alg:struct-sim}
\begin{algorithmic}[1]
\Require GT events $G$, GEN events $H$, novel name $z$, threshold $\tau=0.5$; weights $(\alpha,\beta,\gamma)=(0.6,0.2,0.2)$
\Ensure Structural score $S_{\text{struct}}$, components $(\bar{S}_{\text{event}}, \text{Coverage}, \text{Ordering})$
\State $(\mathcal{A}, \bar{S}_{\text{event}}) \gets \textsc{ThresholdedHungarianAlignment}(G,H,z,\tau)$ \Comment{Alg.~\ref{alg:threshold-hungarian}}
\State $n_g \gets |G|$;\; $n_h \gets |H|$;\; $\text{Coverage} \gets \dfrac{|\mathcal{A}|}{\max(n_g,n_h)}$
\If{$|\mathcal{A}| < 2$}
    \State $\text{Ordering} \gets 0$
\Else
    \State $\mathbf{i} \gets [\,i \mid (i,j)\in \mathcal{A}\,]$;\quad $\mathbf{j} \gets [\,j \mid (i,j)\in \mathcal{A}\,]$
    \State $\tau_K \gets \textsc{KendallTau}(\mathbf{i},\mathbf{j})$ \Comment{SciPy \texttt{kendalltau}}
    \If{$\tau_K$ is NaN} \State $\tau_K \gets 1$ \EndIf
    \State $\text{Ordering} \gets (\tau_K + 1)/2$ \Comment{map from $[-1,1]$ to $[0,1]$}
\EndIf
\State $S_{\text{struct}} \gets \alpha\,\bar{S}_{\text{event}} + \beta\,\text{Coverage} + \gamma\,\text{Ordering}$
\State \Return $S_{\text{struct}}, \bar{S}_{\text{event}}, \text{Coverage}, \text{Ordering}$
\end{algorithmic}
\end{algorithm}

\endgroup

\section{Human Evaluation Survey: Similarity of Narrative Continuations}
\label{appendix:human_eval}
\begin{tcolorbox}[
  breakable,
  colback=white,
  colframe=black,
  boxrule=0.5pt,
  arc=2pt,
  left=8pt,right=8pt,top=8pt,bottom=8pt,
  title={Human Evaluation Packet (per passage pair)}
]

\subsection*{Task Overview}
Annotators are presented with two short passages:
\begin{itemize}
    \item \textbf{Passage A}: human-written continuation (ground truth).
    \item \textbf{Passage B}: model-generated continuation.
\end{itemize}
The task is to compare the two passages and evaluate their similarity along three dimensions:
\emph{linguistic style}, \emph{narrative structure}, and \emph{overall authorial authenticity}.

\medskip
\subsection*{Annotation Guidelines}

\paragraph{Linguistic Style Similarity.}
This dimension captures how similar the two passages \emph{sound} at the sentence and paragraph level.
Annotators should consider:
(1) \textbf{vocabulary choice} (e.g., colloquial vs.\ formal, repeated motifs),
(2) \textbf{sentence structure and rhythm} (e.g., long descriptive clauses vs.\ short direct sentences),
(3) \textbf{tone and mood} (e.g., playful, dark, detached, or emotional), and
(4) \textbf{figurative language or stylistic markers} (e.g., metaphors, imagery, rhetorical devices).
High similarity means that both passages give a comparable stylistic impression regardless of exact wording.

\paragraph{Narrative Structure Similarity.}
This dimension focuses on the organization of the story and progression of events.
Annotators should evaluate:
(1) \textbf{event coverage} — whether both passages describe similar key events or actions,
(2) \textbf{character consistency} — whether the same characters appear and play comparable roles, and
(3) \textbf{ordering} — whether events unfold in a similar sequence or causal chain.
Ratings should reflect the degree to which the generated passage preserves the structural logic of the original continuation.

\paragraph{Overall Authorial Authenticity.}
Provide a holistic judgment of whether the two passages could plausibly have been written by the same author,
integrating evidence from both linguistic style and narrative structure.
This overall score serves as the primary human evaluation of ``same-author likelihood.''

\medskip
\subsection*{Rating Scale}
All dimensions are rated on a 1--5 Likert scale:
\begin{center}
\begin{tabular}{@{}ccccc@{}}
\toprule
\textbf{1} & \textbf{2} & \textbf{3} & \textbf{4} & \textbf{5} \\
Very different & Different & Moderately similar & Similar & Very similar \\
\bottomrule
\end{tabular}
\end{center}

\medskip
\subsection*{Survey Form (per passage pair)}
\textbf{Passage A (Ground Truth)}\\
\textit{[Insert text here]}\\[0.75em]

\textbf{Passage B (Model Output)}\\
\textit{[Insert text here]}\\[1em]

\textbf{Q1. Linguistic Style Similarity (1--5)}\\[0.25em]
$\square$ 1 \quad $\square$ 2 \quad $\square$ 3 \quad $\square$ 4 \quad $\square$ 5 \\[0.75em]

\textbf{Q2. Narrative Structure Similarity (1--5)}\\[0.25em]
$\square$ 1 \quad $\square$ 2 \quad $\square$ 3 \quad $\square$ 4 \quad $\square$ 5 \\[0.75em]

\textbf{Q3. Overall Authorial Authenticity (1--5)}\\[0.25em]
$\square$ 1 \quad $\square$ 2 \quad $\square$ 3 \quad $\square$ 4 \quad $\square$ 5 \\[0.75em]

\textbf{Q4. Brief Justification (Optional)}\\
\textit{[Free text response here]}

\end{tcolorbox}

\section{Sample Ground Truth and Generation Results}
\label{appendix:sample_results}
\begin{mdframed}
\textbf{End of Context of \textit{Never Flinch}:}
\texttt{
 ...Izzy, mindful of Holly Gibney’s pet peeve about insurance companies: “I’m surprised the company didn’t find a way to wiggle out of it. I mean, he did frame a man who got murdered in prison. Did you about know that?” “Of course I know,” the nurse says. “He brags about how sorry he is. Seen a minister. I say crocodile tears!” Tom says, “The DA declined to prosecute, says Tolliver’s full of shit, so he gets a pass and his insurance company gets the bill.” The nurse rolls her eyes. “He’s full of something, all right. Try the solarium first.” As they walk down the corridor, Izzy thinks that if there’s an afterlife, Alan Duffrey may be waiting there for his one-time colleague, Cary Tolliver. “And he’ll want to have a few words.” Tom looks at her. “What?” “Nothing.”}\\
 
 \textbf{Ground Truth of \textit{Never Flinch}:}
 \texttt{Holly pulls the last of the Global Insurance forms in front of her, sighs, grabs her pen—these forms have to be filled out by hand if she wants a chance at finding the missing trinkets, God knows why —and then puts it down. She picks up her phone and looks at the letter from Bill Wilson, whoever he might really be. It’s not her case and she’d never poach it from Isabelle, but Holly can feel her lights turning on, nevertheless. Her job is often boring, there’s too much paperwork, and right now cases—good ones, engaging ones—are thin on the ground, so she’s interested. There’s something else, too, even more important. When her interior lights come on... she loves that. Adores it. “This is not my business. Shoemaker, stick to thy last.” One of her father’s sayings. Her late mother, Charlotte, had a thousand pithy aphorisms, her father only a few... but she remembers every one of them. What is a shoemaker’s last, anyway? She has no idea and quashes the urge to google it. She does know what her last is: filling out this last form, then checking pawnshops and fences for a bunch of jewelry stolen from a rich widow in Sugar Heights. If she can find that stuff, she’ll get a bonus from Buster the Talking Donkey. Which he’ll probably poop out of his butt, she thinks. Very reluctantly. She sighs, picks up her pen again, puts it down, and writes an email instead. Iz—You’ll know this already, it’s pretty obvious, but the guy you are looking for is smart. He talks about the Blackstone Rule, which isn’t in an uneducated man’s vocabulary. I believe the innocent should be punished for the needless death of an innocent might be a cuckoo sentiment, but you have to admit it’s a nicely turned phrase. Balanced. All his punctuation is perfect. Note the use of colons in the heading and how he uses Cc in reference to Chief Patmore. In the old days, when I was doing office correspondence, that stood for “carbon copy.” Now it just means “also sent to,” and is commonly used in business. Suggests to me your Bill Wilson may be a white-collar guy. Now as to that name, Bill Wilson. I don’t think he picked it out of a hat. (Assuming he is male. ) It’s not impossible that he met the murdered man, Alan Duffrey, in AA or NA. (Also assuming it’s Duffrey the letter-writer is on about. )...}\\

\textbf{Generation by \textit{Baseline}:}
\texttt{Izzy, Tom, and Cary Tolliver sit in the solarium, surrounded by plants and sunlight, looking out at the park. Cary Tolliver, a thin, wizened man with gray hair, is lying on a chaise longue, holding a book in his hand. He’s dressed in a pale blue suit that looks like it’s been in the kid’s section of a department store. He doesn’t look up as Izzy and Tom approach. “I see you’ve found me,” he says. “I was hoping you’d come.” He doesn’t sound weak, not yet. “We’re not here to interrogate you,” Izzy says. “Yet. We’re just here to listen. Cary, we know about the... “ “The what?” “The allegations against you, and your claim that you framed Alan Duffrey for kiddie porn. Can you tell us more about that?” Cary Tolliver closes his book, puts it on the table in front of him, and sits up. “That’s a lie. I did frame him, but it was just to save him from the worst of it. He was going to get killed in prison, and I couldn’t bear the thought of that. I wanted to protect him.” “Protect him from what?” “The... “ “The system,” Izzy says gently. “You framed him so he wouldn’t have to go to prison.” “I didn’t mean to hurt anyone. I was just trying to... “ “To what?” “To save him from the worst of it. That’s all I meant.” Izzy looks at Tom, who shrugs. “Tell us about Alan Duffrey’s relationship with the other people in his life,” Izzy says. “His wife and kids? Anyone else? Did he have any enemies or people who might have wanted to hurt him?” Cary Tolliver leans back in his chaise longue, his eyes drifting out the window. “He was married, but they’re divorced. No kids. Just his parents. He had a few friends, but they’re all either dead or in prison.” Izzy raises an eyebrow. “In prison? Why’s that?” “Some of his... associates were involved in some shady dealings. They got pinched. Alan got caught up in it and was eventually sent away.” Cary Tolliver pauses, collecting his thoughts. “His parents were a big part of it. They were involved in some shady business...}\\

\textbf{Generation by \textit{Profile}:}
\texttt{They’re on the fourth floor. The sun is low in the sky, and the light is quite dim. A few steps from the nurses’ station, Tolliver’s door is now in the center of the hall. It’s a bit of a distance from the other doors, so they don’t know who’s there, if there’s any one there. Izzy and Tom have to enter through the back door. They push it open, and the door slams shut behind them, leaving two large, dark, empty spaces. That’s when Tolliver, in the white suit with the khaki slacks, the same suits he always wears, and the same belt buckle, strolled in with a can of Diet Coke in one hand and a shopping bag in the other. He’d left his slacks on the chair in the corner, where he’d been sitting. It’s a bit of a distance from the chair to the door, but both men can walk quickly on a flat surface. Tolliver, with a handbag in each hand—each handbag with a shopping bag in it—had almost no luggage. “It’s a good thing my suit is well-fitted,” the old man says as he comes into a clearing. He’s standing at a half-dozen feet, with arms hanging down. “I wasn’t about to go out without my favorite stuff.” “I know,” Tom says. “We don’t go out without our favorite stuff, either.” Tolliver smiles. “What are you thinking about, Tom?” “How about some of those?” Tom pulls an orange juice out of the bag, and Tolliver takes a swig. “I’ll have to try that, I think. It’s always been a weakness of mine.” “You look nice,” Izzy says. “I mean, not too bad for your age.” “You used to look like a beautiful bird, Holly.” Holly is about to say something, but Tolliver interrupts. “There’s a woman upstairs, and she says it’s time we got on our feet, and it’s really, really late.”...}

\textbf{Generation by \textit{Concept+Linguistic}:}
\texttt{Holly sees her car in the parking lot, but it’s not the Prius she remembers. The rear window is frosted over, which is normal, although the glass is clear on the other side. It looks like it’s been sitting there a while, but not long. The top is down, and the interior is clean except for the usual coffee stains. Holly gets out. She’s surprised to see that the license plate is new, the only change. The car’s been repainted, and now it’s a silver 2015 Ford F150. She walks around it, taking a long look at the rear, and finds the dent she remembers. She looks at the front. The bumper is in the same place, but it’s not dented. It’s been replaced. For the first time, Holly believes the car has been tampered with. “You’re kidding,” she says, turning to go back to her office. The guy who cleans her car for her, a guy named Jack, says, “No ma’am, I’m not,” when she walks back in. “I’m here for my regular cleaning, but they told me you’re sick. I came early, just in case.” Holly smiles. “You’re a good man.” “You’re the best,” Jack says. “Now, where’s my money?” “You’re not going to do it and then change your mind, are you?” “Oh no, ma’am. I’m just the man. I got the car back.” He looks at the rear. “It’s the same as before.” “I see.” “I’d have brought it to you, but the garage woulda charged you for a new bumper.” Jack nods. “So you got it for nothing.” “Well, not really.” “But you needed it less than the garage. Got it.” “Got it.” Jack leans over, and Holly thinks he might be about to kiss her. But he doesn’t. He’s reaching for the key chain. “Hey, it’s the weekend. You’re not working, right?” “No, ma’am. Today’s the day after Labor Day, ma’am. The school’s closed.” “That’s true. Have you been here before?” “No, ma’am, I’m new. To Finders Keepers. I got this job a couple of months ago.” Holly can’t help it; she smiles, and it costs her. “And you’re new to Finders Keepers, too.” “I’m new to every place I go, ma’am.” He gets in the car and starts it. Holly reaches for the passenger door handle, but he’s already gone. “I’ll drive you home,” he says. “Thank you, Jack,” she says, “but you can’t.” He stops the car. “Why not?” “I’m going to see my mother.” “You’re going where?” “Bell, with my mother.” He looks at her. “Where’s she at?” “In Bell. She’s a professor, in the English department.” “And you’re not going to school today?” “No.” “And you’re not going to Bell tonight?” “No. I’m going there tomorrow.” “Fine. So you’re going home now?” “Yes.” “Where?” “Home.” “Now that’s a question.” He has the radio on...}

\textbf{Summary:}\\
\textit{The \textbf{Concept+Linguistic} generation is the closest to the ground truth because it follows Holly’s point of view and mixes everyday details with moments of suspicion, just like the original. It uses clear description and natural dialogue, which feels much closer to the ground truth style than the other generations. The scene also moves at a similar pace, starting from small observations and building toward a sense that something is wrong. This makes it a better match in both writing style and story structure.}

\end{mdframed}
\section{The Use of Large Language Models}

In this work, we used large language models (LLMs) as supportive tools rather than research contributors. Specifically, LLMs assisted with polishing the writing for clarity and readability and with refining prompt designs for our experiments. However, they were not involved in research ideation, conceptual framing, or substantive writing of the paper. All outputs produced with LLM assistance were carefully reviewed and verified by the authors, who take full responsibility for the content of this research.

\end{document}